\documentclass[10pt,twocolumn,letterpaper]{article}
\newcommand\todo[1]{{\color{red}TODO: #1}}

\usepackage{template/CVPR/cvpr}
\usepackage{times}
\usepackage{epsfig}
\usepackage{graphicx}
\usepackage{amsmath}
\usepackage{amssymb}

\usepackage{multirow}

\usepackage{booktabs} 
\usepackage{subcaption}
\usepackage{esint}

\usepackage[pagebackref=true,breaklinks=true,letterpaper=true,colorlinks,bookmarks=false]{hyperref}

\ifcvprfinal\fi

\def\cvprPaperID{10030} 
\cvprfinalcopy

\title{Repurposing GANs for One-shot Semantic Part Segmentation}

\author{Nontawat Tritrong\footnotemark[1] \and Pitchaporn Rewatbowornwong\footnotemark[1] \and Supasorn Suwajanakorn \and
VISTEC, Thailand\\
{\tt\small \{nontawat.t\_s19, pitchaporn.r\_s18, supasorn.s\}@vistec.ac.th}
}
\begin{document}


\twocolumn[{%
\renewcommand\twocolumn[1][]{#1}%
\maketitle

\begin{center}
\centering
  \includegraphics[scale=0.39]{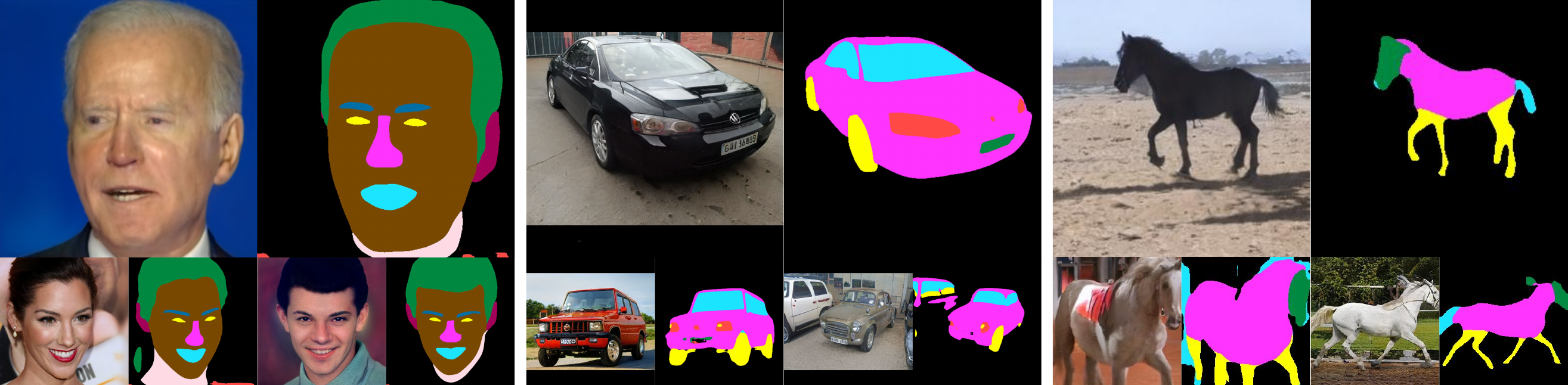}
  \captionof{figure}{One-shot segmentation results. In each task, our segmentation network is given only one example of part labels. }
  \label{fig-intro}
\end{center}
}]

\begin{abstract}
\footnotetext[1]{Authors contributed equally to this work.}
While GANs have shown success in realistic image generation, the idea of using GANs for other tasks unrelated to synthesis is underexplored. Do GANs learn meaningful structural parts of objects during their attempt to reproduce those objects? In this work, we test this hypothesis and propose a simple and effective approach based on GANs for semantic part segmentation that requires as few as one label example along with an unlabeled dataset. 
Our key idea is to leverage a trained GAN to extract a pixel-wise representation from the input image and use it as feature vectors for a segmentation network. Our experiments demonstrate that this GAN-derived representation is ``readily discriminative'' and produces surprisingly good results that are comparable to those from supervised baselines trained with significantly more labels. We believe this novel repurposing of GANs underlies a new class of unsupervised representation learning, which can generalize to many other tasks.
More results are available at \emph{\small \url{https://RepurposeGANs.github.io/}}.

\end{abstract}

\section{Introduction}
After seeing what an elephant trunk looks like for the first time, a young child can identify this conspicuous part for the whole herd. This key capability in humans is still a fundamental challenge in computer vision. That is, how can a machine learn to identify an object or its parts by seeing only one or few examples? A kid does, however, have access to prior visual information learned constantly throughout the years, and he or she could quickly learn to identify human ears perhaps by utilizing the experience of seeing many faces before. In this paper, we tackle a problem inspired by this scenario. Given a large photo collection of human faces, or any other object classes, our goal is  to identify the pixels corresponding to each semantic part for unseen face images given \emph{very few} images with part annotations.

This problem setup is different from the typical definition of few-shot learning, which describes a problem where a learning algorithm trained with many object classes needs to classify or operate on new classes with few supervised examples of those new classes. In contrast, our novel few-shot setup involves a single object class with few annotated examples and no other training data from any other classes. Many methods are proposed in this area of few-shot learning, and the general idea is to apply prior knowledge learned externally to the few-shot task. Examples include meta learning \cite{shaban2017one} and prototype representation \cite{liu2020part, wang2019panet} which extract information from annotations of non-target classes or image-level annotations to be used as prior knowledge. However, 
most of these approaches still learn from some supervised task that requires expensive labels or part annotations. 
In this work, we introduce a new direction that uses a generative model, specifically a generative adversarial network (GAN) ~\cite{goodfellow2014generative}, to learn this prior knowledge from zero labels and apply it to semantic segmentation.

GANs have been highly successful in modeling the data distribution and generating realistic images \cite{karras2019style,karras2020analyzing,brock2018large}. We hypothesize that GANs need to learn meaningful structural information of objects in order to synthesize them correctly, and the generative computations required to synthesize different parts of object could provide useful discriminative information for other tasks \cite{adiwardana2016using, rezagholiradeh2018reg}. Our main contribution is a method that leverages a trained GAN to extract meaningful pixel-wise representations from images. These representations can then be used directly for semantic part segmentation. Our experiments show that GANs are incredibly effective for learning such representations and can achieve surprisingly good segmentation results with only one example label (see Figure \ref{fig-intro}). To our knowledge, this is the first time such high-quality results are achieved on one-shot part segmentation. 

Despite its remarkable results, this core idea alone heavily relies on time-consuming latent optimization and requires the test image to lie close to the image distribution learned by GANs. In this paper, we also demonstrate a simple extension, called auto-shot segmentation, that can bypass the latent optimization, leading to faster and more efficient predictions. And importantly, by performing geometric data augmentation during auto-shot training, we can segment multiple objects with different sizes and orientations all at once---a real-world scenario unseen during training.

To summarize, our main contribution is a novel use of GANs for unsupervised pixel-wise representation learning, which achieves surprising and unprecedented performance on few-shot semantic part segmentation. Our findings reveal that such a representation is readily discriminative. We also demonstrate how to extend the main idea to real-world scenarios to address some of the domain gap between the GAN's training data and real-world images.

  \section{Related Work}

\textbf{Representation Learning}
The goal of representation learning is to capture the underlying information from raw data that is useful and more convenient to process for downstream tasks. Many approaches learn these representations from solving one task and employ them to help improve the performance on another task ~\cite{donahuedeep,sharif2014cnn,girshick2014rich,sermanet2013overfeat,dai2016instance,iglovikov2018ternausnet}. Recent studies have demonstrated that representations learned through a self-supervised task can boost the performance of supervised tasks such as classification and segmentation. Examples of these self-supervision tasks include spatial relative position prediction~\cite{doersch2015unsupervised, noroozi2016unsupervised}, image colorization~\cite{larsson2017colorization}, and image transformation classification~\cite{karras2019style,dosovitskiy2014discriminative}. In contrast, our work explores a representation learned from a generative task, i.e., image synthesis, and extracts feature vectors at the pixel level that is more effective for segmentation problems.

\textbf{Generative Models} Deep generative models have shown promising results in modeling image distribution, thus enabling synthesis of realistic images. There are several classes of visual generative models which include autoregressive models \cite{oord2016pixel,van2016conditional}, autoencoders based on encoder-decoder architectures such as VAE and its variants \cite{kingma2013auto,higgins2016beta, chen2018isolating, vahdat2020nvae}, and 
generative adversarial networks (GANs) \cite{goodfellow2014generative}. Currently, GANs are best in class in image synthesis and have been applied to many other tasks such as image completion \cite{pathakCVPR16context} and  image-to-image translation \cite{pix2pix2017,CycleGAN2017}. State-of-the-art GANs, such as StyleGAN2~\cite{karras2020analyzing} and BigGAN~\cite{brock2018large}, can generate extremely realistic images at high resolution. Our work employs GANs for representation learning, and we provide a study that shows the effectiveness of GANs over alternative models.

Motivated by the impressive results from GANs, numerous studies attempt to understand and interpret the internal representations of GANs. GANs dissection~\cite{bau2018gan} applies an external segmentation model to find the relationship between feature maps and output objects, which also allows adding and removing objects in the output image. Suzuki et al.~\cite{suzuki2018spatially} show that interchanging activations between images can result in interchanging of objects in the output image. Edo et al.~\cite{collins2020editing} use clustering to find distinctive groups of feature maps and allow spatially localized part editing. Tsutsui et al. \cite{tsutsui2019meta} improve one-shot image recognition by combining images synthesized by GANs with the original training images. \cite{donahue2019large} uses representations learned from BigGAN and achieves state-of-the-art performance on unsupervised representation learning on ImageNet~\cite{russakovsky2015imagenet}. 

Some other studies analyze GANs through manipulation of the latent code and attempt to make GANs' internal representations more interpretable.
Chen et al.\cite{chen2016infogan} use mutual information to force the network to store human-interpretable attributes in their latent code. Shu et al.\cite{shu2017neural} solve a similar problem via an additional encoder network, which allows users to have control over the generated results.
AttGAN \cite{he2019attgan} exploits an external classifier network to enable attribute editing. Voynov $\&$ Babenko \cite{voynov2020unsupervised} propose an unsupervised method to discover interpretable directions in the latent space.  
In this paper, we leverage the insight that GANs internal representations are tightly coupled to the generated output and that they can hold useful semantic information. 

\indent\textbf{Semantic Part Segmentation}
Semantic part segmentation aims to segment parts within an object as opposed to objects within a scene as in semantic segmentation. This problem can be more challenging because two parts sometimes do not have a visible boundary between them, such as nose and face. Considerable progress has been made in semantic part segmentation \cite{wang2015semantic, wang2015joint, tsogkas2015deep, naha2020pose}, but these techniques demand a vast number of pixel-wise annotations.

To avoid using pixel-wise annotations, some approaches rely instead on other kinds of annotations that are cheaper to obtain, such as keypoints \cite{fang2018weakly}, body poses \cite{yang2019weakly}, or edge maps \cite{zhang2020correlating}. However, they are often inflexible and only work on some specific domains, such as human body parts. Some other attempts forgo the annotations completely with self-supervised techniques. For example, \cite{hung2019scops} uses equivariance, geometric, and semantic consistency constraints to train a segmentation network, and \cite{lathuiliere2020motion, xu2019unsupervised} exploit motion information from videos. One main drawback of these unsupervised methods is that there is little control over the partition of object parts, which can lead to arbitrary segmentation. Unlike these approaches, our method allows complete control over the partition of object parts by requiring only few annotated examples.

\indent\textbf{Few-shot Semantic Segmentation} Past research has attempted to solve segmentation with few annotations. A meta learning approach \cite{shaban2017one} first trains a segmentation network on an annotated dataset then fine-tunes the network parameters on one annotation of the target class. Prototypical methods \cite{dong2018few, liu2020part, wang2019panet} use a support set to learn a prototype vector for each object class.
Both meta learning and prototypical methods construct two training branches where the support branch is trained on annotations of non-target classes or image-level annotations, and the query branch then takes an input image as well as the extracted feature to predict segmentation masks. Similarity guidance network \cite{zhang2020sg} masks off the background in the support image, then finds the pixels in the query branch with similar foreground features. Some work \cite{siam2019one, caelles2017one} segment objects in all video frames with only the first frame annotated. Nonetheless, these methods have not shown success in semantic part segmentation. Meta learning requires annotation masks of similar object classes, and hence learning part-specific prototypes is not viable. Leveraging the information from the support set is also difficult due to the lack of part-level annotations. In contrast, our representation extracted from GANs contains part-level information and can be learned without supervision.
  \begin{figure}
\centering
  \includegraphics[scale=0.23]{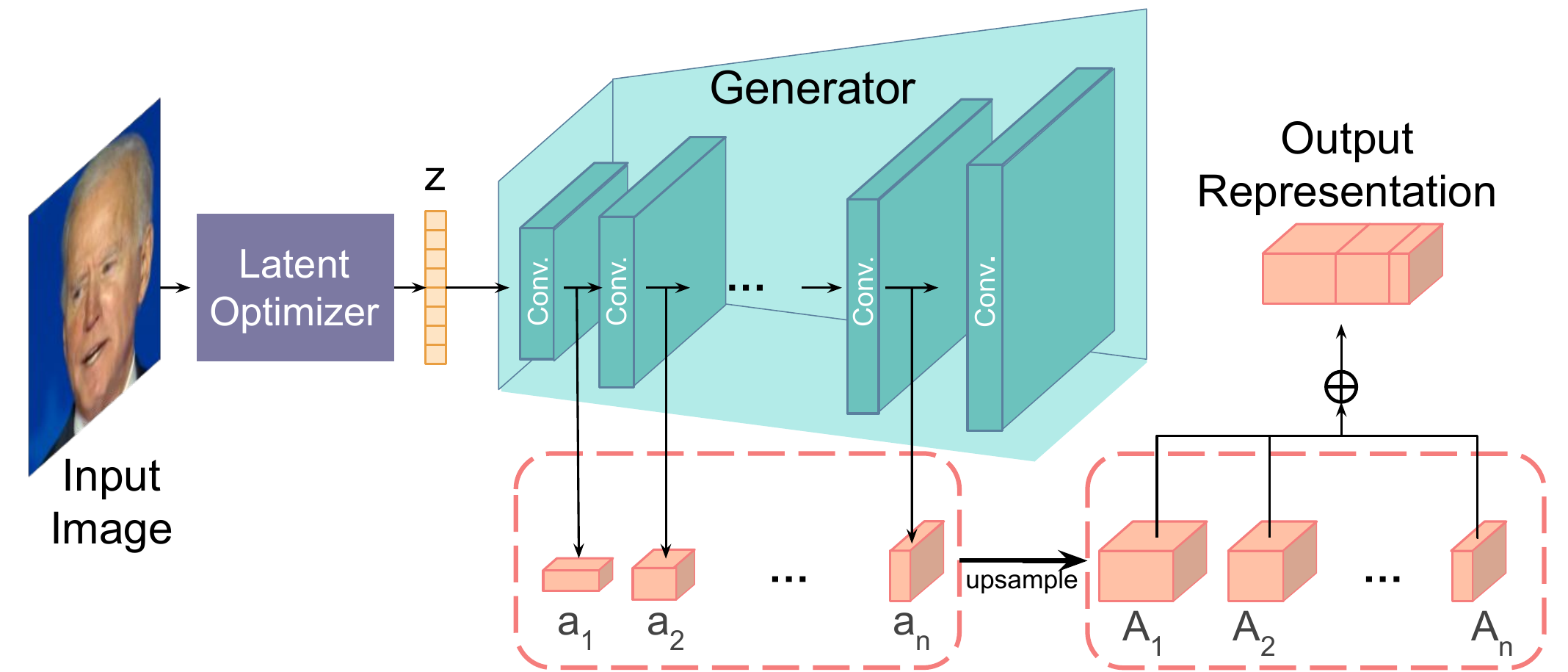}
  \caption{\textbf{Representation extraction} To extract a representation from an image, we embed the image into the latent space of GAN by optimizing for the latent $z$ that reproduces the input image. $z$ is then fed to the generator and we collect multiple activation maps $a_{1}, a_{2}, ..., a_{n}$ of dimensions $(h_1, w_1, c_1), ..., (h_n, w_n, c_n)$. Each of these maps is upsampled to $A_i$ with dimension $(h_n, w_n, c_i)$. The representation is a concatenation of all $A_{i}$ along the channel dimension.
  } 
  \label{pipe-feature-extract}
\end{figure}

\begin{figure}
\centering
  \includegraphics[scale=0.31]{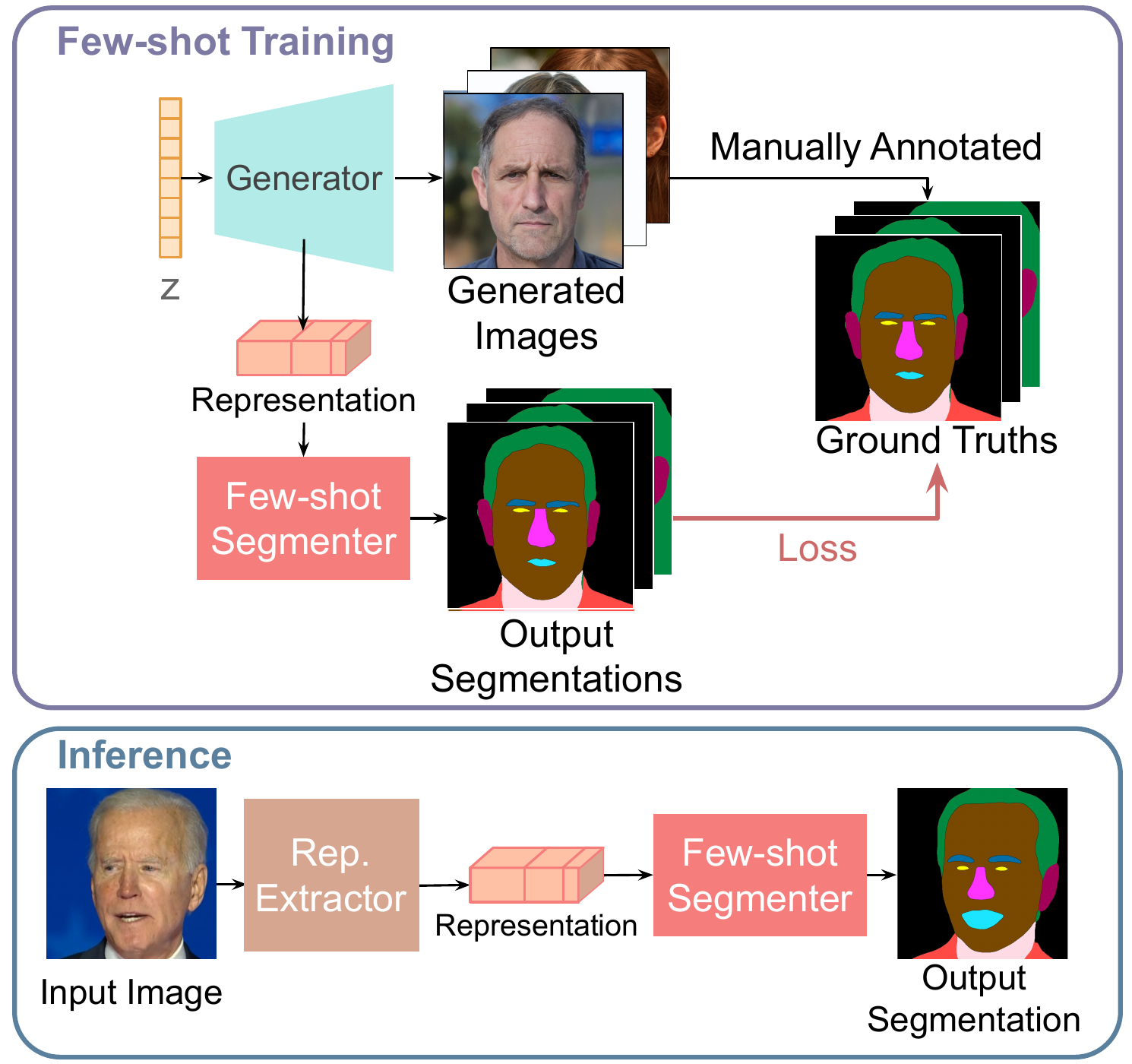}
  \caption{\textbf{Few-shot segmentation pipeline} For training, we use a trained GAN to generate a few images along with their representations by feeding random latent codes. Then, we manually annotate these images and train our few-shot segmenter to output segmentation maps that match our annotated masks. For inference, we extract a representation from a test image (Figure \ref{pipe-feature-extract}) then input it to the few-shot segmenter to obtain a segmentation map. 
  }
  \label{pipe-primary}
  \vspace{-1.5em}
\end{figure}

\begin{figure}
\centering
  \includegraphics[scale=0.35]{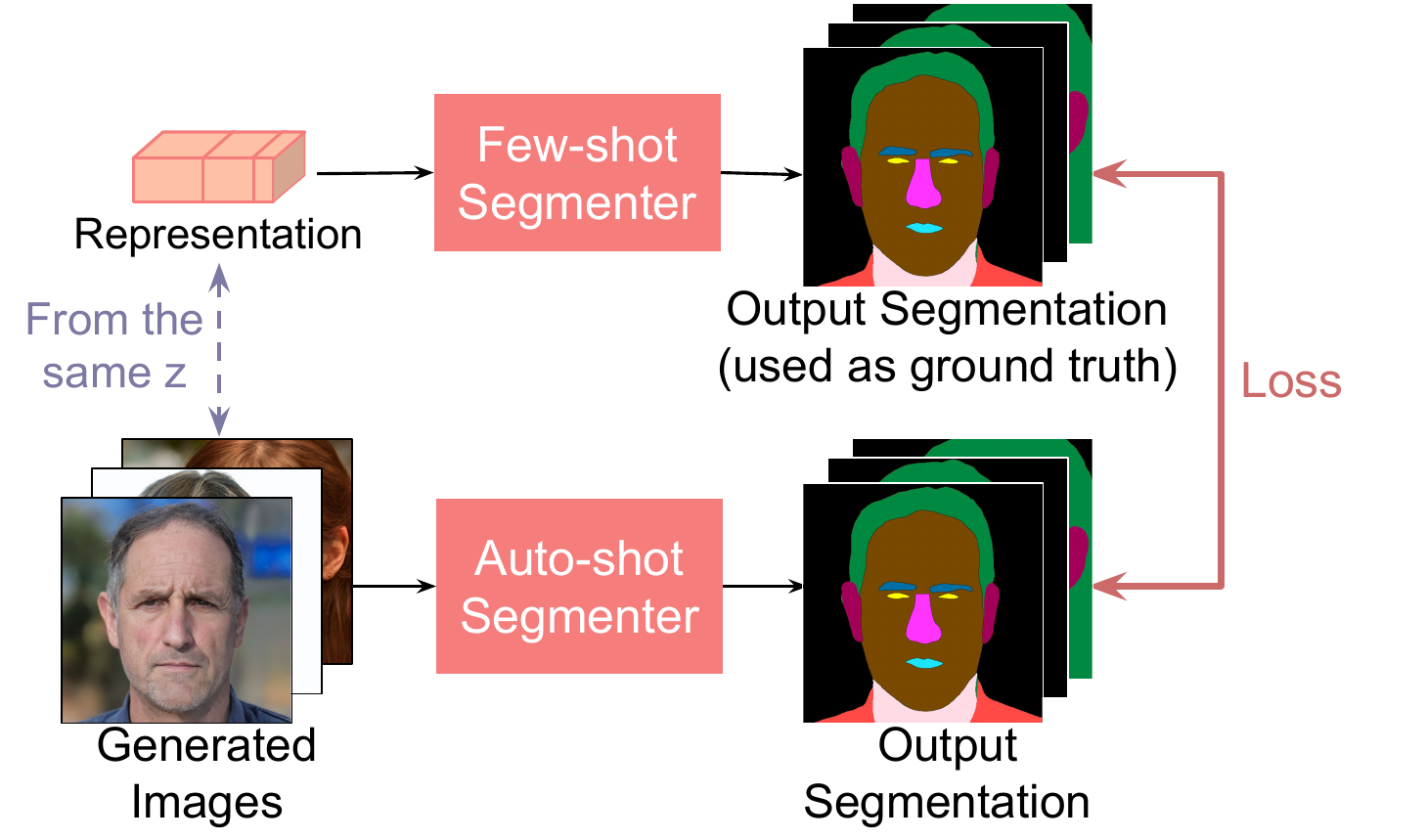}
  \caption{\textbf{Auto-shot segmentation pipeline} during training, the auto-shot segmenter uses generated images from the trained GAN as input and segmentation masks predicted by the few-shot segmenter as ground truth.}
  \label{pipe-secondary}
\end{figure}

\section{Approach}
Our problem concerns semantic part segmentation with the following novel setup. Given a set of unlabeled images and a few images (1-10) with part annotations from a single object class, our goal is to part-segment an unseen object from the same class. These part annotations can be specified by the user with binary masks. 
Note that semantic part segmentation can also be considered as an $n$-way \emph{per-pixel} classification problem where $n$ is the number of parts.

This problem would become trivial if there existed a function $f$ that maps each pixel value, which by itself lacks semantic meaning, to its own feature vector that contains discriminative information for part classification. We propose to derive such a function from a GAN trained to synthesize images of the target class. In the following sections, we will explain how a GAN is utilized for this task, how to use the computed per-pixel features for segmentation, and finally a simple extension that allows segmentation without requiring a GAN or its expensive mapping during inference.

\subsection{Representation Extraction from GANs}\label{repextract}
Using GANs as a mapping function is not straightforward simply because GANs take as input a random latent code, not the image pixels to be mapped. To understand our process, first consider a typical scenario where we generate an image by feeding a random latent code to a convolutional-based GAN. In this case, the synthesized output image is constructed by the generator through a series of spatial convolutions, and each output pixel is a result of a unique generative computation that can be traced back through each convolution layer down to the initial latent code. 

Our key idea is to use these unique computational ``paths'' for feature representation. Generally, the computational path for generating a pixel is a directed acyclic graph with nodes representing the network parameters or the input latent code involved in the computation of that pixel. However, in our work these nodes represent activation values, and we simply represent the path with a single sequence of activations from all layers within the generator that are spatially aligned with that pixel. In particular, as shown in Figure \ref{pipe-feature-extract},
we extract the activation map from every layer (or some subset of layers) of the generator, $a_1, a_2, \ldots, a_n$, each with dimension $(h_i, w_i, c_i)$, and compute our pixel-wise representation as
\begin{equation}F = \mathbb{U}(a_{1})\oplus_{c}\mathbb{U}(a_{2})\oplus_{c}...\oplus_{c}\mathbb{U}(a_{n})\end{equation} where $\mathbb{U}(\cdot)$ spatially upsamples the input to the size of the largest activation map $(h_n, w_n)$ and $\oplus_{c}$ is a concatenation along the channel dimension. This process maps each 3-dimensional RGB pixel to a $C-$dimensional feature vector, where $C=\sum_{i=1}^n c_i$.

Normally, this extraction process only works for images that are synthesized by the generator and cannot be used directly for real test images. However, given any test image, one can optimize for a latent code that generates that given test image with any gradient-based optimization or with more sophisticated schemes \cite{karras2020analyzing, abdal2019image2stylegan, gabbay2019style}. The resulting latent code then allows the feature map to be constructed in a similar manner.

\subsection{Segmentation with Extracted Representation}
To solve few-shot segmentation, we first train a GAN on images of our target class and generate $k$ random images by feeding it random latent codes. Then, we compute the feature maps and manually annotate object parts for these $k$ images. The $k$ feature maps and annotations together form our supervised training pairs which can be used to train a segmentation model, such as a multilayer perceptron or a convolutional network (see Figure \ref{pipe-primary}). To segment a test image, we compute a pixel-wise feature map for the test image using the aforementioned latent code optimization and feed it to our trained segmentation network. 

\subsection{Extension: Auto-shot Segmentation Network}
Computing pixel-wise feature vectors using a GAN does have a number of restrictions. First, the test image needs to lie close to the image distribution modeled by the GAN; otherwise, the latent optimization may fail to reproduce the test image, leading to poor feature vectors. This constrain can be severely restricting for classes like human face because the input image has to contain exactly one face aligned and centered similar to the trainset. Second, relying on a GAN to generate feature vectors through the latent optimization process is expensive and time-consuming as it requires multiple forward-backward passes through the generator.

To overcome these limitations, we use our trained GAN to synthesize a large set of images and predict segmentation maps for those images using our network to form paired training data. To retain all the probability information from our network's prediction process, each pixel in our segmentation maps is represented by a set of logit values of all part labels rather than a single part ID. That is, we do not apply softmax and argmax to produce the segmentation maps.
With this training data, we train another network, e.g. a UNet \cite{ronneberger2015u}, to solve the segmentation in a single network pass from raw images without relying on a GAN or its feature mapping (see Figure \ref{pipe-secondary}).
We call this process auto-shot segmentation. Additionally, we apply data augmentation that allows detection of objects at different scales and orientations. Interestingly, we demonstrate how this simple approach can successfully segment multiple object instances at the same time with good quality in our experiments and supplementary video. 

  \section{Implementation}
Our training pipeline starts by training a GAN on a dataset of the target class. Then, we use the trained GAN to generate a few images along with their pixel-wise representations (Section \ref{repextract}) and manually annotate these images with the desired part segmentation. Finally, we train a few-shot segmentation network that takes as input the pixel-wise representation to predict an output segmentation. For the auto-shot segmentation, we use the same GAN to generate a large dataset of images and use our trained few-shot network to predict segmentation maps for those images. These generated images and their corresponding segmentation maps are then used to train the auto-shot segmentation network. 

\subsection{Generative Adversarial Network}
We use StyleGAN2 \cite{karras2020analyzing} for our pipeline. StyleGAN2 has 9 pairs of convolution layers with activation outputs of sizes: $4^2$, $8^2$, $16^2$, $32^2$, $64^2$, $128^2$, $256^2$, $512^2$, and $1024^2$. For feature extraction, we use all pairs except the last which generates the output image. We also use StyleGAN2's projection method proposed in their paper to embed an image into the latent space (latent code optimization explained in Section \ref{repextract}). 


\subsection{Few-shot Segmentation Network} The few-shot network takes the $C$-channel pixel-wise representation as input and outputs a segmentation map. We explore 2 different architectures: fully convolutional networks (CNN) and multilayer perceptrons (MLP).

\textbf{CNN:} We first use a linear embedding layer (1x1 convolution) to reduce the input dimension from $C$ to 128, followed by 8 convolutional layers with a kernel size of 3 and dilation rates of 2, 4, 8, 1, 2, 4, 8, and 1. The dimensions of the output channels are: 64 for the first 6 layers, 32, and the number of classes. All layers except the output layer use leaky ReLU activation functions.

\textbf{MLP:} We use a 2-layer MLP with 2,000 and 200 hidden nodes for the first and second layers. All layers except the output layer use ReLU activation functions.

Both MLP and CNN were trained for 1,000 epochs with a cross-entropy loss and a weight decay of 0.001 using Adam optimizer. Our initial learning rate is 0.001 with a decay factor of 0.9 every 50 epochs.

\subsection{Auto-shot Segmentation Network} 
This network is trained with GAN's generated images and their corresponding segmentation maps from the few-shot network. We adopt a UNet architecture described in our supplementary. Additionally, we perform the following data augmentation on this training set: 1) random horizontal flips, 2) random scales between 0.5 and 2, 3) random rotations between -10 and 10 degree, 4) random vertical and horizontal translations between 0\% and 50\% of the image size.
This network is trained for 300 epochs using Adam optimizer with an initial learning rate of 0.001 and a decay factor of 0.1 when the validation score does not decrease within 20 epochs.
  
\begin{table}[]
\centering
\caption{Weighted IOU scores on few-shot human face segmentation.}
\label{tab:GAN-face-perclass}
\small
\begin{tabular}{ccc|l|l|cll}
\bottomrule
Segmentation Network & Shots & \multicolumn{3}{c}{3-class}     & \multicolumn{3}{c}{10-class}    \\ \bottomrule
        & 1     & \multicolumn{3}{c}{71.7}          & \multicolumn{3}{c}{77.9}          \\
              {CNN}  & 5     & \multicolumn{3}{c}{82.1}          & \multicolumn{3}{c}{83.9}          \\
                     & 10    & \multicolumn{3}{c}{\textbf{83.5}} & \multicolumn{3}{c}{\textbf{85.2}} \\ \hline
           & 1     & \multicolumn{3}{c}{75.3}          & \multicolumn{3}{c}{74.1}          \\
              {MLP}  & 5     & \multicolumn{3}{c}{77.8}          & \multicolumn{3}{c}{79.6}          \\
                     & 10    & \multicolumn{3}{c}{77.2}          & \multicolumn{3}{c}{77.2}          \\ \bottomrule
\end{tabular}
\end{table}

\begin{figure}
\centering
  \includegraphics[scale=0.56]{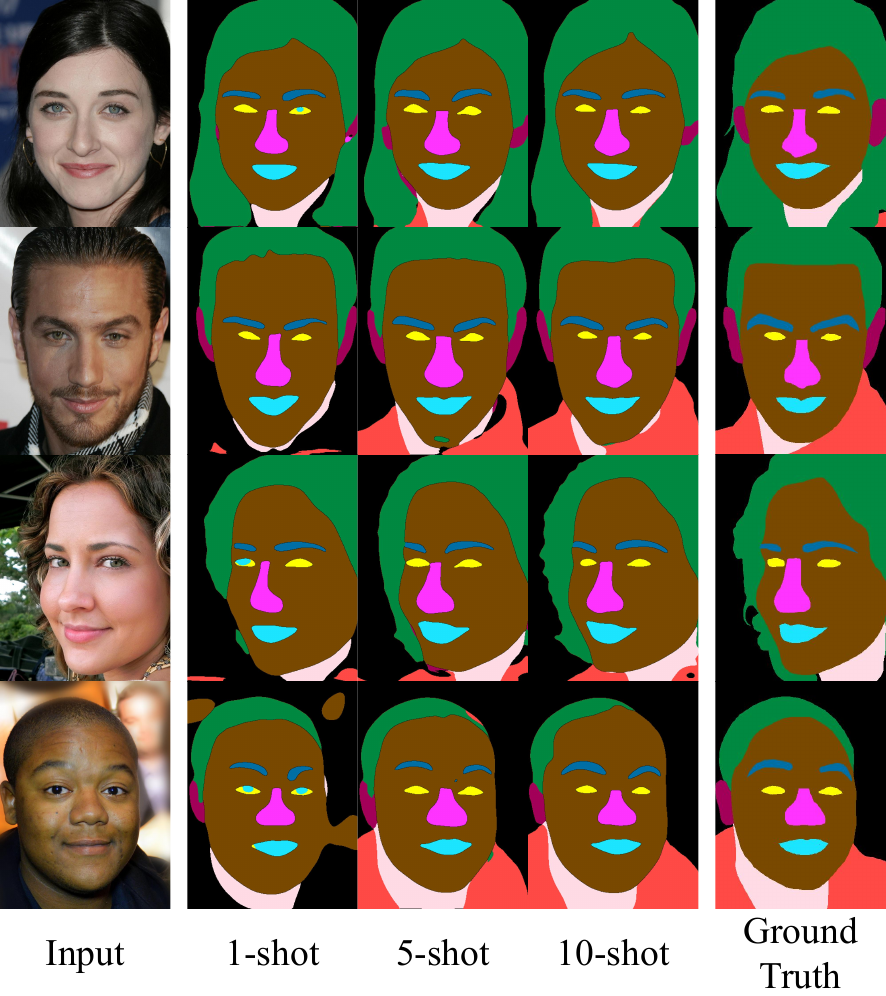}
  \caption{Few-shot face segmentation results on CelebAMASK-HQ.}
  \label{fig-gan-face}
  \vspace{-1.5em}
\end{figure}

\begin{figure}
\centering
  \includegraphics[scale=0.5]{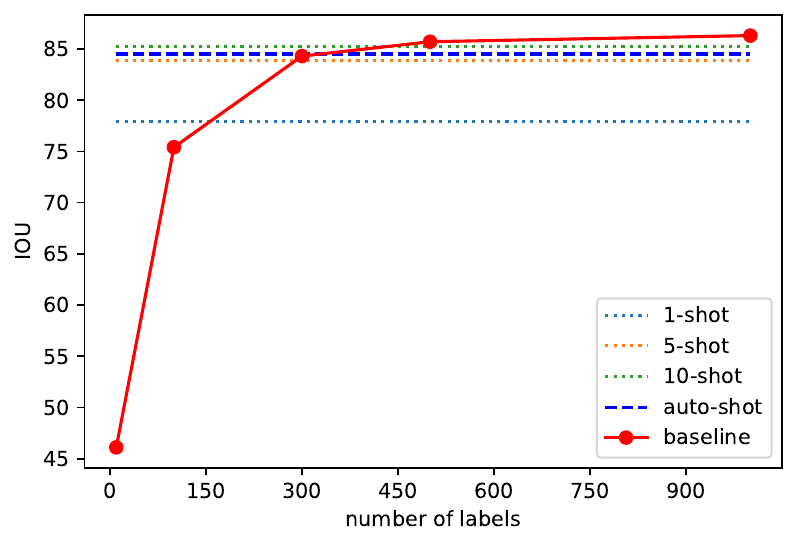}
  \caption{10-class face segmentation results of supervised baseline and the number of segmentation labels used. Our few-shot segmentation results are shown in dot line for comparison. Supervised baseline consumes over 100 annotations to surpass our 1-shot segmenter, and around 500 annotation to reach same-level of IOU on our 10-shot segmenter.
 }
  \label{fig-GAN-baseline}
  \vspace{-0.7em}
\end{figure}

\begin{table*}[]
\centering
\caption{IOU scores of our 10-shot vs auto-shot segmenters on 10-class face segmentation. The auto-shot segmenter is trained with a dataset generated by the 10-shot segmenter. Both techniques have similar performance, which demonstrates the effectiveness of the dataset generation and auto-shot training process.}
\label{tab:few-vs-auto}
\resizebox{\columnwidth*2}{!}{%
\begin{tabular}{cccccccccccc}
\bottomrule
Network               & Weighted IOU & Eyes & Mouth & Nose & Face & Clothes & Hair  & Eyebrows & Ears & Neck & BG   \\ \hline
10-shot segmenter  & 85.2          & 74.0 & 84.6  & 82.9 & 90.0    & 23.6    & 79.2  & 63.1    & 27.0 & 73.6 & 84.2 \\
Auto-shot segmenter & 84.5          & 75.4 & 86.5  & 84.6 & 90.0    & 15.5    & 84.0  & 68.2    & 37.3 & 72.8 & 84.7 \\ \bottomrule
\vspace{-2em}
\end{tabular}}
\end{table*}

\begin{table}[]
\centering
\caption{Per-class IOU scores on 3-class human face segmentation.}
\label{tab:GAN-face}
\small
\begin{tabular}{ccccc}
\bottomrule
                              & Weighted IOU & Eyes  & Mouth & Nose\\ \hline
 \multicolumn{1}{c}{1-shot}  & 71.7                  & 57.8 & 71.1  & 76.0\\
 \multicolumn{1}{c}{5-shot}  & 82.1                  & 73.6 & 84.0  & 82.1\\ 
\multicolumn{1}{c}{10-shot} & 83.5                  & 75.9 & 85.3  & 82.7 \\\bottomrule
\vspace{-1.5em}
\end{tabular}
\end{table}

\begin{figure*}
\centering
  \includegraphics[scale=0.56]{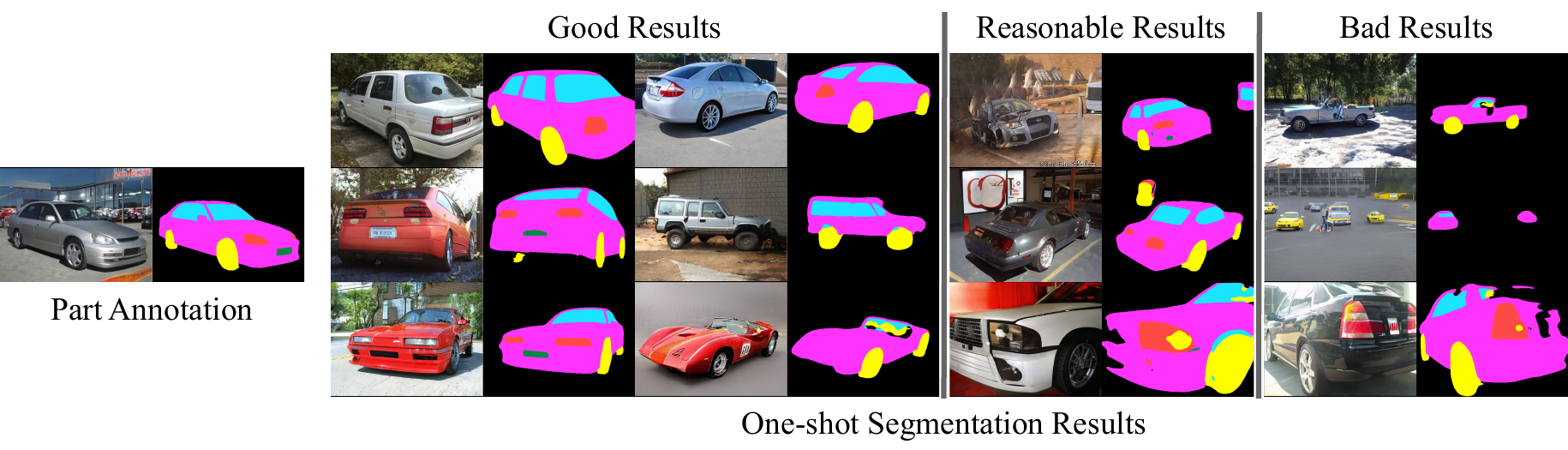}
  \caption{One-shot car part segmentation results on GAN's generated images. The segmentation network can segment car images from varied points of view even though it is trained on annotations of one car from one angle. However, there are some failure cases when the cars appear unusually big or small, or when GANs generate unrealistic cars.}
  \label{fig-car-1shot}
\end{figure*}


\begin{table}[]
\caption{IOU scores on PASCAL-Parts car segmentation.}
\label{tab:Pas-car-bg}
\resizebox{\columnwidth}{!}{%
\begin{tabular}{cccccccc}
\bottomrule
Model         & Body & Plate & Light & Wheel & Window & BG   & Average \\ \hline
CNN\cite{tsogkas2015deep}           & 73.4 & 41.7  & 42.2  & 66.3  & 61      & 67.4 & 58.7    \\ 
CNN+CRF\cite{tsogkas2015deep}       & 75.4 & 35.8  & 36.1  & 64.3  & 61.8    & 68.7 & 57      \\ 
Ours (Auto-shot)           & 75.5 & 17.8  & 29.3  & 57.2  & 62.4    & 70.7 & 52.2    \\ \bottomrule
OMPS\cite{zhao2019ordinal}  & 86.3 & 50.5  & 55.1 & 75.5 & 65.2 &-   & 66.5   \\
Ours (Auto-shot) w/o bg   & 76.4 & 17.5 & 29.3 & 52.5 & 64.1    & -     & 47.9   \\ \bottomrule
\end{tabular}}
\end{table}

\begin{table}[]
\caption{IOU scores on PASCAL-Parts horse segmentation. ``-" indicates no available result.}
\label{tab:Pas-horse-part}
\resizebox{\columnwidth}{!}{%
\begin{tabular}{cccccccc}
\bottomrule
Model            & Head  & Neck & Torso & Neck+Torso & Legs  & Tail  & BG    \\ \hline
Shape+Appearance\cite{wang2015semantic} & 47.2 & -    & -     & 66.7      & 38.2 & -     & -     \\ 
CNN+CRF\cite{tsogkas2015deep}          & 55.0  & 34.2 & 52.4  & -          & 46.8  & 37.2  & 76.0    \\
Ours (Auto-shot)         & 50.1 & -    & -     & 70.5      & 49.6 & 19.9 & 81.6 \\ \bottomrule
\vspace{-2em}
\end{tabular}}
\end{table}

\begin{table}[]
\centering
\caption{Average IOU scores on PASCAL-Parts horse segmentation.}
\resizebox{\columnwidth}{!}{%
\label{tab:Pas-horse}
\begin{tabular}{cccc}
\bottomrule
\multicolumn{1}{c}{Model} & \multicolumn{1}{c}{RefineNet \cite{fang2018weakly}} & \multicolumn{1}{c}{Pose-Guided \cite{naha2020pose}} & Ours(Auto-shot)   \\ \hline
Horse IOU                     & 36.9          & 60.2            & 53.1 \\ \bottomrule
\end{tabular}}
\end{table}

\begin{figure}
  \includegraphics[scale=0.56]{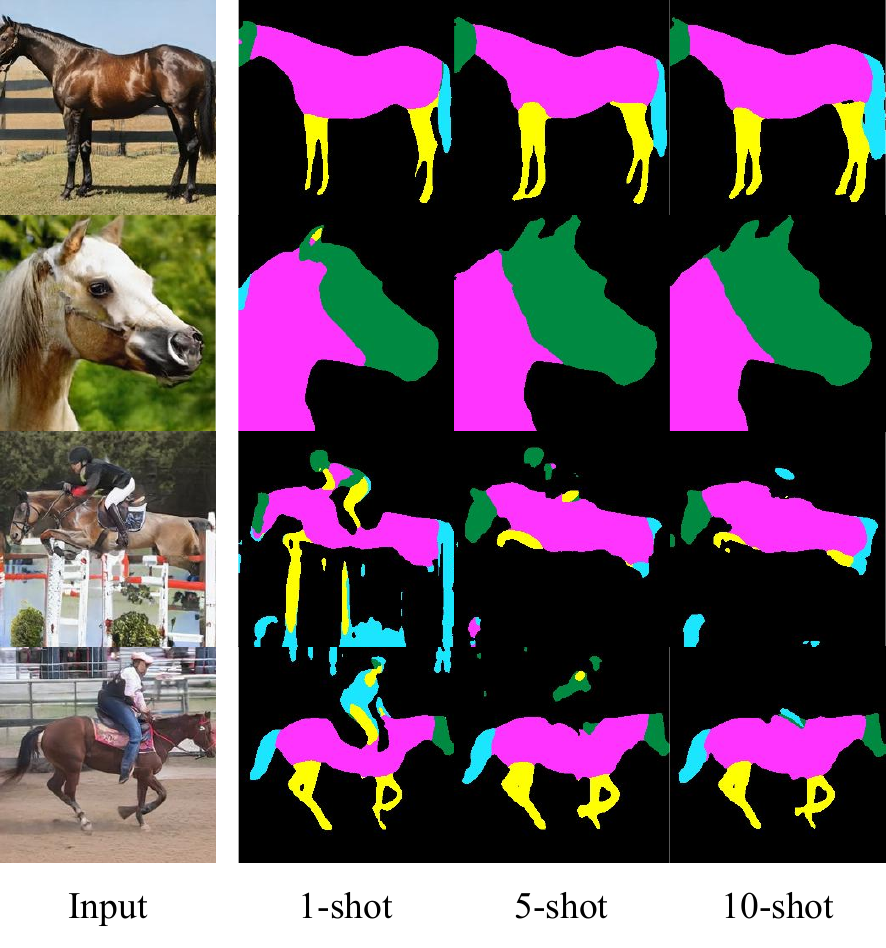}
  \caption{Results on few-shot horse part segmentation from GAN's generated images. Compared to cars and faces with good 1-shot results, horses need more labels. 1-shot horse segmentation often mistakes the rider as a part of horse.}
  \label{fig-gan-horse}
  \vspace{-1.5em}
\end{figure}

\begin{figure*}
\centering
  \includegraphics[scale=0.35]{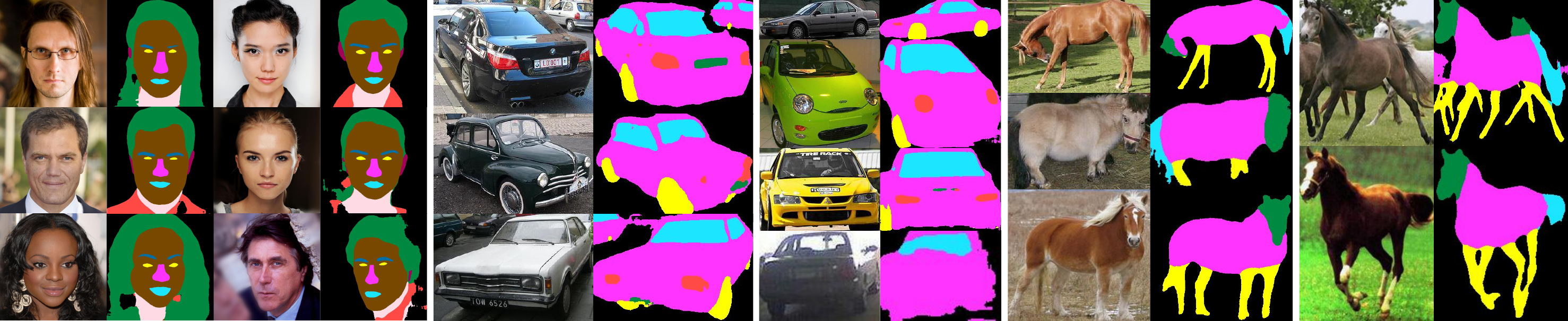}
  \caption{Some examples of auto-shot segmentation trained with datasets generated by 10-shot segmenters on CelebAMask-HQ, PASCAL-Part Car and PASCAL-Part horse. The pretrained StyleGAN2 for each class was trained on FFHQ, LSUN-Car and LSUN-Horse.}
  \label{fig-auto-shot}
  \vspace{-1.5em}
\end{figure*}

\section{Experiments}
We perform the following experiments in this section: 1) we evaluate the performance of our few-shot and auto-shot segmenters on 3 object classes and compare them to baselines, 2) we evaluate alternative structures of the few-shot segmentation network. Additionally in our supplementary material, 3) we show segmentation results on videos, 4) we study whether the choice of layers used for feature extraction affects the segmentation performance, 5) we test whether our method can segment parts that have arbitrary or unusual shapes that do not correspond to any semantic parts, 6) we explore and evaluate features learned from other generative models, such as VAE, or from other supervised and self-supervised learning methods and compare against our features learned from GANs.


\subsection{Semantic Part Segmentation}
We evaluate segmentation performance of the few-shot and auto-shot segmenters on 3 object classes: human face, car, and horse.



\textbf{Datasets:} To train the few-shot segmenter, we use face images and annotated segmentation masks from CelebAMask-HQ \cite{CelebAMask-HQ}. For horse and car, we use images generated by pretrained StyleGAN2s and manually annotate them ourselves. For the auto-shot segmenter, we use 5,000 images generated from each GAN trained on each object class and the predicted annotations from the few-shot segmenter. Our models are evaluated on CelebAMask-HQ for faces and PASCAL-Part dataset \cite{chen2014detect} for car and horse.

\textbf{Evaluation metric:} We use intersect-over-union (IOU) to evaluate individual object parts and report weighted IOU scores, where the weight of each class is the ratio of the number of ground-truth pixels belonging to the class to the total number of pixels.

\subsubsection{Human Face Part Segmentation}
We perform experiments with 12 combinations of settings that vary 1) the architecture of the few-shot segmenter (CNN or MLP), 2) the number of part classes (3 or 10), and 3) the number of examples with part annotations (1-shot, 5-shot, 10-shot). One interesting finding in Table \ref{tab:GAN-face-perclass} is that the MLP segmenter, which can only look at the features of individual pixels to make per-pixel predictions, performs well and almost similarly to the CNN segmenter that has a wide receptive field and can exploit structure priors for predicting a segmentation map. 

Table \ref{tab:few-vs-auto} shows IOU scores of the 10-shot segmenter and auto-shot segmenter on 10-class face segmentation. Surprisingly, the auto-shot segmenter achieves similar IOU scores to those of 10-shot segmenter in all classes except for clothes, even though it relies only on the dataset generated by the 10-shot segmenter. Note that the few-shot network also performs poorly on clothes relative to other parts, which could be due to the large variation in clothing. The auto-shot segmenter can also segment unaligned images at various scales due to the data augmentation during training. Figure \ref{fig-GAN-baseline} shows a 10-class segmentation comparison between our few-shot and auto-shot segmenters and a supervised baseline which uses the same architecture as the auto-shot segmenter and is trained on ground-truth masks from CelebAMask-HQ with varying numbers of labels. Our few-shot segmenter trained with a single label produces a comparable IOU score to the supervised baseline trained with about 150 labels. And with 10 labels, both of our segmenters match the baseline performance with 500 labels. Qualitative and quantitative results for the CNN-based few-shot network are shown in Figure \ref{fig-gan-face} and Table \ref{tab:GAN-face}. 

\subsubsection{Car Part Segmentation}
Unlike well-aligned face images in CelebA-HQ, car images in PASCAL-Part have larger variations in pose and appearance. Despite this challenge, our one-shot segmenter produces good segmentation results and can identify wheels, windows, and the license plate shown in Figure \ref{fig-car-1shot}. We compare our method to DeepCNN-DenseCRF \cite{tsogkas2015deep} and the Ordinal Multitask Part Segmentation \cite{zhao2019ordinal} on the car class in PASCAL-Part. Details on the experiment setup can be found in our supplementary material. 
Table \ref{tab:Pas-car-bg} shows our results using the auto-shot segmenter trained on a 10-shot dataset (dataset generated from our few-shot segmenter with 10 labels), which compares favorably to the fully supervised baselines. Note that we compare to \cite{zhao2019ordinal} by excluding the background class similarly to how their scores were reported.

\subsubsection{Horse Part Segmentation}
Horse segmentation is more challenging than the other two because horses are non-rigid and can appear in many poses such as standing or jumping. Also, the boundaries between legs and body are not clearly visible. Our one-shot segmenter has lower performance compared to those of faces and cars; however, the result improves significantly with a few more annotations as shown in Figure \ref{fig-gan-horse}. In Table \ref{tab:Pas-horse-part}, we compare our auto-shot segmentation (also learned from dataset by 10-shot segmenter) IOU scores on each class to Shape+Appearance \cite{wang2015semantic} and CNN+CRF \cite{tsogkas2015deep}. 
Table \ref{tab:Pas-horse} shows the overall IOU scores of our auto-shot segmenter, RefineNet \cite{fang2018weakly}, and Pose-Guided Knowledge Transfer \cite{naha2020pose}. The score of RefineNet is taken from \cite{naha2020pose}. Our method surpasses RefineNet, and our IOU is slightly lower than Pose-Guided Knowledge Transfer \cite{naha2020pose} which is a fully-supervised method trained with over 300 annotated images and additional annotated keypoints. Experimental details can be found in our supplementary material. Auto-shot segmentation results are presented in Figure \ref{fig-auto-shot}. 

\begin{table}[]
\centering
\caption{IOU scores on 1-shot face segmentation of different size of few-shot segmenters.}

\label{tab:GAN-exp}
\begin{tabular}{c|c|c}
\bottomrule
Model                & Size            & Weighted IOU \\ \bottomrule
\multirow{3}{*}{MLP} & 0 hidden layers  & 74.0          \\
                     & 1 hidden layer  & 72.2          \\
                     & 2 hidden layers & 74.1          \\ \hline
\multirow{3}{*}{CNN} & S               & 73.4          \\
                     & M               & 75.2          \\
                     & L               & 77.9          \\ \bottomrule
\end{tabular}
\end{table}

\begin{figure}
\centering
  \includegraphics[scale=0.56]{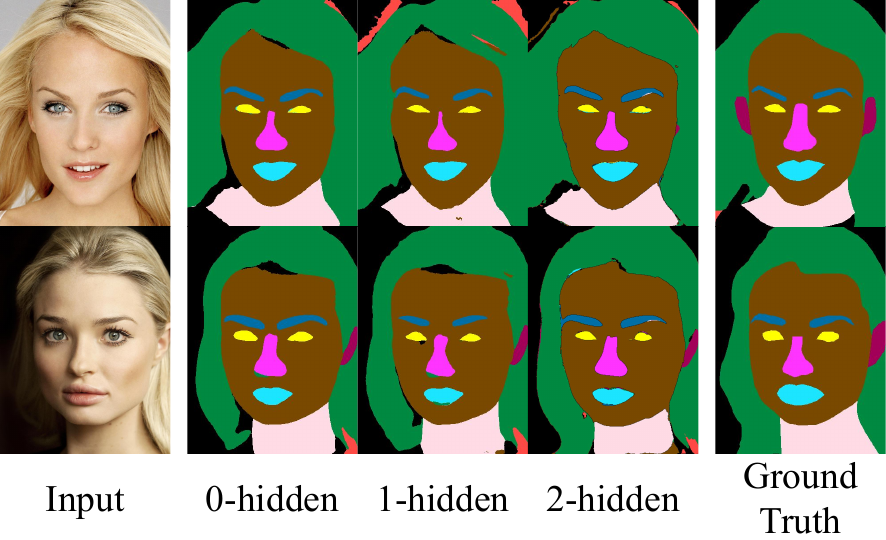}
  \caption{Results on 1-shot segmentation of MLP-based segmentors containing 0, 1, or 2 hidden layers.}
  \label{fig-clus}
  \vspace{-2em}
\end{figure}

\subsection{Analysis on GAN-derived Representation}
One desirable property of a good representation is that it should contain meaningful semantic information in a readily discriminative form. 
We could test this by evaluating how well a simple linear classifier or smaller networks with limited capability perform given our representation as input.

We evaluate several architectures on one-shot face segmentation: 3 sizes of multilayer perceptrons: i) 0 hidden layers, ii) 1 hidden layer with 2000 nodes, and iii) 2 hidden layers with 2000 and 200 nodes, as well as small, medium, and large convolutional networks described in the supplementary material (Table C - E).
We found that a linear classifier (0 hidden-layer) gives reasonable segmentation masks shown in Figure \ref{fig-clus}; however, a non-linear MLP classifier (2 hidden layers) is needed to obtain more accurate boundaries in complex areas such as hair. As shown in Table \ref{tab:GAN-exp}, L-network obtains the highest IOU score, although smaller networks or even a linear classifier do not perform significantly worse with IOU differences of only around 2.7-5.7.

\section{Conclusion}
We have presented a simple and powerful approach that repurposes GANs, used predominantly for synthesis, for few-shot semantic part segmentation. Our novelty lies in the unconventional use of \emph{readily discriminative} pixel-wise representation extracted from the generative processes of GANs.
Our approach achieves promising and unprecedented performance that allows part segmentation given very few annotations and is competitive with fully-supervised baselines that require 10-50$\times$ more label examples. 
We also propose a more efficient extension to our segmentation pipeline that bypasses the required latent optimization and generalizes better to real-world scenarios with multiple objects of varying sizes and orientations. We believe this novel use of GANs for unsupervised representation learning can serve as an effective and generic ``upstream'' task in transfer learning for problems that involve reasoning about object parts, scene semantics, or make pixel-level predictions. 

\section*{Acknowledgements}  
    This research was supported by PTT public company limited and SCB public company limited. 
  {
    \small
    \bibliographystyle{template/CVPR/ieee_fullname}
    \bibliography{content/bibliography.bib}

\begin{thebibliography}{10}\itemsep=-1pt

\bibitem{abdal2019image2stylegan}
Rameen Abdal, Yipeng Qin, and Peter Wonka.
\newblock Image2stylegan: How to embed images into the stylegan latent space?
\newblock In {\em Proceedings of the IEEE international conference on computer
  vision}, pages 4432--4441, 2019.

\bibitem{adiwardana2016using}
Daniel De~Freitas Adiwardana, Akihiro Matsukawa, and Jay Whang.
\newblock Using generative models for semi-supervised learning.
\newblock In {\em Medical image computing and computer-assisted
  intervention--MICCAI}, volume 2016, pages 106--14, 2016.

\bibitem{bau2018gan}
David Bau, Jun-Yan Zhu, Hendrik Strobelt, Bolei Zhou, Joshua~B Tenenbaum,
  William~T Freeman, and Antonio Torralba.
\newblock Gan dissection: Visualizing and understanding generative adversarial
  networks.
\newblock {\em arXiv preprint arXiv:1811.10597}, 2018.

\bibitem{brock2018large}
Andrew Brock, Jeff Donahue, and Karen Simonyan.
\newblock Large scale gan training for high fidelity natural image synthesis.
\newblock {\em arXiv preprint arXiv:1809.11096}, 2018.

\bibitem{caelles2017one}
Sergi Caelles, Kevis-Kokitsi Maninis, Jordi Pont-Tuset, Laura Leal-Taix{\'e},
  Daniel Cremers, and Luc Van~Gool.
\newblock One-shot video object segmentation.
\newblock In {\em Proceedings of the IEEE conference on computer vision and
  pattern recognition}, pages 221--230, 2017.

\bibitem{chen2018isolating}
Ricky~TQ Chen, Xuechen Li, Roger~B Grosse, and David~K Duvenaud.
\newblock Isolating sources of disentanglement in variational autoencoders.
\newblock In {\em Advances in Neural Information Processing Systems}, pages
  2610--2620, 2018.

\bibitem{chen2016infogan}
Xi Chen, Yan Duan, Rein Houthooft, John Schulman, Ilya Sutskever, and Pieter
  Abbeel.
\newblock Infogan: Interpretable representation learning by information
  maximizing generative adversarial nets.
\newblock In {\em Advances in neural information processing systems}, pages
  2172--2180, 2016.

\bibitem{chen2014detect}
Xianjie Chen, Roozbeh Mottaghi, Xiaobai Liu, Sanja Fidler, Raquel Urtasun, and
  Alan Yuille.
\newblock Detect what you can: Detecting and representing objects using
  holistic models and body parts.
\newblock In {\em Proceedings of the IEEE conference on computer vision and
  pattern recognition}, pages 1971--1978, 2014.

\bibitem{collins2020editing}
Edo Collins, Raja Bala, Bob Price, and Sabine Susstrunk.
\newblock Editing in style: Uncovering the local semantics of gans.
\newblock In {\em Proceedings of the IEEE/CVF Conference on Computer Vision and
  Pattern Recognition}, pages 5771--5780, 2020.

\bibitem{dai2016instance}
Jifeng Dai, Kaiming He, and Jian Sun.
\newblock Instance-aware semantic segmentation via multi-task network cascades.
\newblock In {\em Proceedings of the IEEE Conference on Computer Vision and
  Pattern Recognition}, pages 3150--3158, 2016.

\bibitem{doersch2015unsupervised}
Carl Doersch, Abhinav Gupta, and Alexei~A Efros.
\newblock Unsupervised visual representation learning by context prediction.
\newblock In {\em Proceedings of the IEEE international conference on computer
  vision}, pages 1422--1430, 2015.

\bibitem{donahuedeep}
Jeff Donahue, Yangqing Jia, Oriol Vinyals, Judy Hoffman, Ning Zhang, Eric
  Tzeng, and Trevor Darrell.
\newblock A deep convolutional activation feature for generic visual
  recognition.
\newblock {\em UC Berkeley \&amp; ICSI, Berkeley, CA, USA}.

\bibitem{donahue2019large}
Jeff Donahue and Karen Simonyan.
\newblock Large scale adversarial representation learning.
\newblock In {\em Advances in Neural Information Processing Systems}, pages
  10542--10552, 2019.

\bibitem{dong2018few}
Nanqing Dong and Eric~P Xing.
\newblock Few-shot semantic segmentation with prototype learning.
\newblock In {\em BMVC}, volume~3, 2018.

\bibitem{dosovitskiy2014discriminative}
Alexey Dosovitskiy, Jost~Tobias Springenberg, Martin Riedmiller, and Thomas
  Brox.
\newblock Discriminative unsupervised feature learning with convolutional
  neural networks.
\newblock In {\em Advances in neural information processing systems}, pages
  766--774, 2014.

\bibitem{fang2018weakly}
Hao-Shu Fang, Guansong Lu, Xiaolin Fang, Jianwen Xie, Yu-Wing Tai, and Cewu Lu.
\newblock Weakly and semi supervised human body part parsing via pose-guided
  knowledge transfer.
\newblock {\em arXiv preprint arXiv:1805.04310}, 2018.

\bibitem{gabbay2019style}
Aviv Gabbay and Yedid Hoshen.
\newblock Style generator inversion for image enhancement and animation.
\newblock {\em arXiv preprint arXiv:1906.11880}, 2019.

\bibitem{girshick2014rich}
Ross Girshick, Jeff Donahue, Trevor Darrell, and Jitendra Malik.
\newblock Rich feature hierarchies for accurate object detection and semantic
  segmentation.
\newblock In {\em Proceedings of the IEEE conference on computer vision and
  pattern recognition}, pages 580--587, 2014.

\bibitem{goodfellow2014generative}
Ian Goodfellow, Jean Pouget-Abadie, Mehdi Mirza, Bing Xu, David Warde-Farley,
  Sherjil Ozair, Aaron Courville, and Yoshua Bengio.
\newblock Generative adversarial nets.
\newblock In {\em Advances in neural information processing systems}, pages
  2672--2680, 2014.

\bibitem{he2016deep}
Kaiming He, Xiangyu Zhang, Shaoqing Ren, and Jian Sun.
\newblock Deep residual learning for image recognition.
\newblock In {\em Proceedings of the IEEE conference on computer vision and
  pattern recognition}, pages 770--778, 2016.

\bibitem{he2019attgan}
Zhenliang He, Wangmeng Zuo, Meina Kan, Shiguang Shan, and Xilin Chen.
\newblock Attgan: Facial attribute editing by only changing what you want.
\newblock {\em IEEE Transactions on Image Processing}, 28(11):5464--5478, 2019.

\bibitem{higgins2016beta}
Irina Higgins, Loic Matthey, Arka Pal, Christopher Burgess, Xavier Glorot,
  Matthew Botvinick, Shakir Mohamed, and Alexander Lerchner.
\newblock beta-vae: Learning basic visual concepts with a constrained
  variational framework.
\newblock 2016.

\bibitem{hung2019scops}
Wei-Chih Hung, Varun Jampani, Sifei Liu, Pavlo Molchanov, Ming-Hsuan Yang, and
  Jan Kautz.
\newblock Scops: Self-supervised co-part segmentation.
\newblock In {\em Proceedings of the IEEE Conference on Computer Vision and
  Pattern Recognition}, pages 869--878, 2019.

\bibitem{iglovikov2018ternausnet}
Vladimir Iglovikov and Alexey Shvets.
\newblock Ternausnet: U-net with vgg11 encoder pre-trained on imagenet for
  image segmentation.
\newblock {\em arXiv preprint arXiv:1801.05746}, 2018.

\bibitem{pix2pix2017}
Phillip Isola, Jun-Yan Zhu, Tinghui Zhou, and Alexei~A Efros.
\newblock Image-to-image translation with conditional adversarial networks.
\newblock {\em CVPR}, 2017.

\bibitem{karras2019style}
Tero Karras, Samuli Laine, and Timo Aila.
\newblock A style-based generator architecture for generative adversarial
  networks.
\newblock In {\em Proceedings of the IEEE conference on computer vision and
  pattern recognition}, pages 4401--4410, 2019.

\bibitem{karras2020analyzing}
Tero Karras, Samuli Laine, Miika Aittala, Janne Hellsten, Jaakko Lehtinen, and
  Timo Aila.
\newblock Analyzing and improving the image quality of stylegan.
\newblock In {\em Proceedings of the IEEE/CVF Conference on Computer Vision and
  Pattern Recognition}, pages 8110--8119, 2020.

\bibitem{kingma2013auto}
Diederik~P Kingma and Max Welling.
\newblock Auto-encoding variational bayes.
\newblock {\em arXiv preprint arXiv:1312.6114}, 2013.

\bibitem{larsson2017colorization}
Gustav Larsson, Michael Maire, and Gregory Shakhnarovich.
\newblock Colorization as a proxy task for visual understanding.
\newblock In {\em Proceedings of the IEEE Conference on Computer Vision and
  Pattern Recognition}, pages 6874--6883, 2017.

\bibitem{lathuiliere2020motion}
St{\'e}phane Lathuili{\`e}re, Sergey Tulyakov, Elisa Ricci, Nicu Sebe, et~al.
\newblock Motion-supervised co-part segmentation.
\newblock {\em arXiv preprint arXiv:2004.03234}, 2020.

\bibitem{CelebAMask-HQ}
Cheng-Han Lee, Ziwei Liu, Lingyun Wu, and Ping Luo.
\newblock Maskgan: Towards diverse and interactive facial image manipulation.
\newblock In {\em IEEE Conference on Computer Vision and Pattern Recognition
  (CVPR)}, 2020.

\bibitem{liu2020part}
Yongfei Liu, Xiangyi Zhang, Songyang Zhang, and Xuming He.
\newblock Part-aware prototype network for few-shot semantic segmentation.
\newblock {\em arXiv preprint arXiv:2007.06309}, 2020.

\bibitem{liu2015faceattributes}
Ziwei Liu, Ping Luo, Xiaogang Wang, and Xiaoou Tang.
\newblock Deep learning face attributes in the wild.
\newblock In {\em Proceedings of International Conference on Computer Vision
  (ICCV)}, December 2015.

\bibitem{naha2020pose}
Shujon Naha, Qingyang Xiao, Prianka Banik, Md Alimoor~Reza, and David~J
  Crandall.
\newblock Pose-guided knowledge transfer for object part segmentation.
\newblock In {\em Proceedings of the IEEE/CVF Conference on Computer Vision and
  Pattern Recognition Workshops}, pages 906--907, 2020.

\bibitem{pytorch-hed}
Simon Niklaus.
\newblock A reimplementation of {HED} using {PyTorch}.
\newblock \url{https://github.com/sniklaus/pytorch-hed}, 2018.

\bibitem{noroozi2016unsupervised}
Mehdi Noroozi and Paolo Favaro.
\newblock Unsupervised learning of visual representations by solving jigsaw
  puzzles.
\newblock In {\em European Conference on Computer Vision}, pages 69--84.
  Springer, 2016.

\bibitem{oord2016pixel}
Aaron van~den Oord, Nal Kalchbrenner, and Koray Kavukcuoglu.
\newblock Pixel recurrent neural networks.
\newblock {\em arXiv preprint arXiv:1601.06759}, 2016.

\bibitem{pathakCVPR16context}
Deepak Pathak, Philipp Kr\"ahenb\"uhl, Jeff Donahue, Trevor Darrell, and Alexei
  Efros.
\newblock Context encoders: Feature learning by inpainting.
\newblock In {\em Computer Vision and Pattern Recognition ({CVPR})}, 2016.

\bibitem{rezagholiradeh2018reg}
Mehdi Rezagholiradeh and Md~Akmal Haidar.
\newblock Reg-gan: Semi-supervised learning based on generative adversarial
  networks for regression.
\newblock In {\em 2018 IEEE International Conference on Acoustics, Speech and
  Signal Processing (ICASSP)}, pages 2806--2810. IEEE, 2018.

\bibitem{ronneberger2015u}
Olaf Ronneberger, Philipp Fischer, and Thomas Brox.
\newblock U-net: Convolutional networks for biomedical image segmentation.
\newblock In {\em International Conference on Medical image computing and
  computer-assisted intervention}, pages 234--241. Springer, 2015.

\bibitem{russakovsky2015imagenet}
Olga Russakovsky, Jia Deng, Hao Su, Jonathan Krause, Sanjeev Satheesh, Sean Ma,
  Zhiheng Huang, Andrej Karpathy, Aditya Khosla, Michael Bernstein, et~al.
\newblock Imagenet large scale visual recognition challenge.
\newblock {\em International journal of computer vision}, 115(3):211--252,
  2015.

\bibitem{sermanet2013overfeat}
Pierre Sermanet, David Eigen, Xiang Zhang, Micha{\"e}l Mathieu, Rob Fergus, and
  Yann LeCun.
\newblock Overfeat: Integrated recognition, localization and detection using
  convolutional networks.
\newblock {\em arXiv preprint arXiv:1312.6229}, 2013.

\bibitem{shaban2017one}
Amirreza Shaban, Shray Bansal, Zhen Liu, Irfan Essa, and Byron Boots.
\newblock One-shot learning for semantic segmentation.
\newblock {\em arXiv preprint arXiv:1709.03410}, 2017.

\bibitem{sharif2014cnn}
Ali Sharif~Razavian, Hossein Azizpour, Josephine Sullivan, and Stefan Carlsson.
\newblock Cnn features off-the-shelf: an astounding baseline for recognition.
\newblock In {\em Proceedings of the IEEE conference on computer vision and
  pattern recognition workshops}, pages 806--813, 2014.

\bibitem{shu2017neural}
Zhixin Shu, Ersin Yumer, Sunil Hadap, Kalyan Sunkavalli, Eli Shechtman, and
  Dimitris Samaras.
\newblock Neural face editing with intrinsic image disentangling.
\newblock In {\em Proceedings of the IEEE conference on computer vision and
  pattern recognition}, pages 5541--5550, 2017.

\bibitem{siam2019one}
Mennatullah Siam, Naren Doraiswamy, Boris~N Oreshkin, Hengshuai Yao, and Martin
  Jagersand.
\newblock One-shot weakly supervised video object segmentation.
\newblock {\em arXiv preprint arXiv:1912.08936}, 2019.

\bibitem{suzuki2018spatially}
Ryohei Suzuki, Masanori Koyama, Takeru Miyato, Taizan Yonetsuji, and Huachun
  Zhu.
\newblock Spatially controllable image synthesis with internal representation
  collaging.
\newblock {\em arXiv preprint arXiv:1811.10153}, 2018.

\bibitem{tsogkas2015deep}
Stavros Tsogkas, Iasonas Kokkinos, George Papandreou, and Andrea Vedaldi.
\newblock Deep learning for semantic part segmentation with high-level
  guidance.
\newblock {\em arXiv preprint arXiv:1505.02438}, 2015.

\bibitem{tsutsui2019meta}
Satoshi Tsutsui, Yanwei Fu, and David Crandall.
\newblock Meta-reinforced synthetic data for one-shot fine-grained visual
  recognition.
\newblock In {\em Advances in Neural Information Processing Systems}, pages
  3063--3072, 2019.

\bibitem{vahdat2020nvae}
Arash Vahdat and Jan Kautz.
\newblock Nvae: A deep hierarchical variational autoencoder.
\newblock {\em arXiv preprint arXiv:2007.03898}, 2020.

\bibitem{van2016conditional}
Aaron Van~den Oord, Nal Kalchbrenner, Lasse Espeholt, Oriol Vinyals, Alex
  Graves, et~al.
\newblock Conditional image generation with pixelcnn decoders.
\newblock In {\em Advances in neural information processing systems}, pages
  4790--4798, 2016.

\bibitem{voynov2020unsupervised}
Andrey Voynov and Artem Babenko.
\newblock Unsupervised discovery of interpretable directions in the gan latent
  space.
\newblock In {\em International Conference on Machine Learning}, pages
  9786--9796. PMLR, 2020.

\bibitem{wang2015semantic}
Jianyu Wang and Alan~L Yuille.
\newblock Semantic part segmentation using compositional model combining shape
  and appearance.
\newblock In {\em Proceedings of the IEEE conference on computer vision and
  pattern recognition}, pages 1788--1797, 2015.

\bibitem{wang2019panet}
Kaixin Wang, Jun~Hao Liew, Yingtian Zou, Daquan Zhou, and Jiashi Feng.
\newblock Panet: Few-shot image semantic segmentation with prototype alignment.
\newblock In {\em Proceedings of the IEEE International Conference on Computer
  Vision}, pages 9197--9206, 2019.

\bibitem{wang2015joint}
Peng Wang, Xiaohui Shen, Zhe Lin, Scott Cohen, Brian Price, and Alan~L Yuille.
\newblock Joint object and part segmentation using deep learned potentials.
\newblock In {\em Proceedings of the IEEE International Conference on Computer
  Vision}, pages 1573--1581, 2015.

\bibitem{xie2015holistically}
Saining Xie and Zhuowen Tu.
\newblock Holistically-nested edge detection.
\newblock In {\em Proceedings of the IEEE international conference on computer
  vision}, pages 1395--1403, 2015.

\bibitem{xu2019unsupervised}
Zhenjia Xu, Zhijian Liu, Chen Sun, Kevin Murphy, William~T Freeman, Joshua~B
  Tenenbaum, and Jiajun Wu.
\newblock Unsupervised discovery of parts, structure, and dynamics.
\newblock {\em arXiv preprint arXiv:1903.05136}, 2019.

\bibitem{yang2019weakly}
Zhengyuan Yang, Yuncheng Li, Linjie Yang, Ning Zhang, and Jiebo Luo.
\newblock Weakly supervised body part parsing with pose based part priors.
\newblock {\em arXiv preprint arXiv:1907.13051}, 2019.

\bibitem{zhang2018unreasonable}
Richard Zhang, Phillip Isola, Alexei~A Efros, Eli Shechtman, and Oliver Wang.
\newblock The unreasonable effectiveness of deep features as a perceptual
  metric.
\newblock In {\em Proceedings of the IEEE conference on computer vision and
  pattern recognition}, pages 586--595, 2018.

\bibitem{zhang2020sg}
Xiaolin Zhang, Yunchao Wei, Yi Yang, and Thomas~S Huang.
\newblock Sg-one: Similarity guidance network for one-shot semantic
  segmentation.
\newblock {\em IEEE Transactions on Cybernetics}, 2020.

\bibitem{zhang2020correlating}
Ziwei Zhang, Chi Su, Liang Zheng, and Xiaodong Xie.
\newblock Correlating edge, pose with parsing.
\newblock In {\em Proceedings of the IEEE/CVF Conference on Computer Vision and
  Pattern Recognition}, pages 8900--8909, 2020.

\bibitem{zhao2019ordinal}
Yifan Zhao, Jia Li, Yu Zhang, Yafei Song, and Yonghong Tian.
\newblock Ordinal multi-task part segmentation with recurrent prior generation.
\newblock {\em IEEE Transactions on Pattern Analysis and Machine Intelligence},
  2019.

\bibitem{CycleGAN2017}
Jun-Yan Zhu, Taesung Park, Phillip Isola, and Alexei~A Efros.
\newblock Unpaired image-to-image translation using cycle-consistent
  adversarial networks.
\newblock In {\em Computer Vision (ICCV), 2017 IEEE International Conference
  on}, 2017.

\end{thebibliography}
  }
  \appendix
    \newpage
\setcounter{table}{0}
\renewcommand{\thetable}{\Alph{table}}
\renewcommand{\thefigure}{\Alph{figure}}
\renewcommand{\thesubsection}{\Alph{subsection}}

\section*{Supplementary Material}
\subsection{Overview}
We present implementation setups and additional experiments in this supplementary material. We clarify the architecture of the auto-shot segmenter in Section B. In Section C, we describe experiment details of car and horse part segmentation mentioned in the main paper. Section D compares using logit values vs one-hot vectors as the ground-truth target labels for the dataset generated by few-shot segmenters. Section E describes architectures used in GAN-derived representation analysis (Section 5.2). Section F investigates the segmentation performance using different choices of GAN's layers. Section G tests our method's ability to segment arbitrary parts. Section H explores representation learned from other unsupervised and self-supervised learning methods and compare their few-shot segmentation performance.

\subsection{Auto-shot segmentor architecture}
We adopt UNet architecture \cite{ronneberger2015u} for our auto-shot segmentation network. The overall architecture is shown in Table \ref{tab:arch}. The network consists of encoder and decoder. For the encoder, there are 5 blocks of a convolutional layer, batch normalization, and a ReLU activation. Max-pooling is used after every 2 blocks to halve the input size. The decoder consists of 4 blocks of bilinear upsampling, 2 convolutional layers, batch normalization, and a ReLU activation. Lastly, a 1x1 convolutional layer is used to map the feature to the segmentation output. 

\subsection{Experiment Setups}
For our experiments in the paper, we try to match the setups of those baseline methods as much as possible for a fair comparison. Most prior part segmentation methods \cite{tsogkas2015deep, zhao2019ordinal, naha2020pose} use bounding boxes to crop the image as they want to focus on only part segmentation not object localization. Some work \cite{naha2020pose, zhao2019ordinal} eliminate objects deemed too small or objects with occlusion. In this section, we clarify the pre-processing step we use for each dataset.

\indent\textbf{Car part segmentation}
We follow the setup from \cite{tsogkas2015deep} and evaluate on PASCAL-Part. We use the provided bounding boxes with class annotations in PASCAL-Part to select and crop images of cars, then use them as our test images.
To compare with \cite{zhao2019ordinal}, we discard images whose bounding boxes overlap with other bounding boxes with IOU more than 5 and images that are smaller than 50x50 pixels. We fill the background with black color. Even though our network still predicts the background class, we calculate the average score without the background class. 

\indent\textbf{Horse part segmentation}
We follow the horse part segmentation's setup from \cite{naha2020pose} and use the provided bounding boxes with class annotations in PASCAL-Part to extract horse images. We discard horse images smaller than 32x32 pixels.

\subsection{Logit vs One-hot Labels for Auto-shot Segmentation}

As explained in the main paper, we use our few-shot segmenter along with a trained GAN to generate a labeled dataset for our auto-shot segmenter. Each training example in this dataset consists of a generated image and its corresponding segmentation map predicted from our few-shot segmenter. For this dataset, each pixel in each segmentation map is represented as a set of logit values corresponding to the probabilities of different part classes, as opposed to a standard one-hot encoding of the part class. The motivation is to keep the class confidence scores that could provide useful information for the auto-shot segmenter and help prevent over-confident predictions based on spurious or ambiguous target labels.

In this experiment, we compare our auto-shot segmenter trained with our proposed logit values to the standard one-hot target labels. Table \ref{tab:logit} shows that using logit labels outperforms one-hot labels on all three object categories. 


\begin{table}[]
\caption{Architecture of auto-shot segmentation network.}

\label{tab:arch}
\resizebox{\columnwidth}{!}{%
\begin{tabular}{|c|c|c|c|c|c|}
\bottomrule
Layer                                                             & Kernel size & Stride & Batch normalization & Activation & Output size       \\ \bottomrule
Input                                                             & -           & -      & No                  & -          & H x W x 3         \\ \hline
Conv1a                                                            & 3 x 3       & 1      & Yes                 & ReLU       & H x W x 64        \\ \hline
Conv1b                                                            & 3 x 3       & 1      & Yes                 & ReLU       & H x W x 64        \\ \hline
Max Pool                                                          & 2 x 2       & 2      & No                  & -          & H/2 x W/2 x 64    \\ \hline
Conv2a                                                            & 3 x 3       & 1      & Yes                 & ReLU       & H/2 x W/2 x 128   \\ \hline
Conv2b                                                            & 3 x 3       & 1      & Yes                 & ReLU       & H/2 x W/2 x 128   \\ \hline
Max Pool                                                          & 2 x 2       & 2      & No                  & -          & H/4 x W/4 x 128   \\ \hline
Conv3a                                                            & 3 x 3       & 1      & Yes                 & ReLU       & H/4 x W/4 x 256   \\ \hline
Conv3b                                                            & 3 x 3       & 1      & Yes                 & ReLU       & H/4 x W/4 x 256   \\ \hline
Max Pool                                                          & 2 x 2       & 2      & No                  & -          & H/8 x W/8 x 256   \\ \hline
Conv4a                                                            & 3 x 3       & 1      & Yes                 & ReLU       & H/8 x W/8 x 512   \\ \hline
Conv4b                                                            & 3 x 3       & 1      & Yes                 & ReLU       & H/8 x W/8 x 512   \\ \hline
Max Pool                                                          & 2 x 2       & 2      & No                  & -          & H/16 x W/16 x 512 \\ \hline
Conv5a                                                            & 3 x 3       & 1      & Yes                 & ReLU       & H/16 x W/16 x 512 \\ \hline
Conv5b                                                            & 3 x 3       & 1      & Yes                 & ReLU       & H/16 x W/16 x 512 \\ \hline
\begin{tabular}[c]{@{}c@{}}Upsample\\ Concat(Conv4b)\end{tabular} & -           & -      & No                  & -          & H/8 x W/8 x 1024  \\ \hline
Conv6a                                                            & 3 x 3       & 1      & Yes                 & ReLU       & H/8 x W/8 x 512   \\ \hline
Conv6b                                                            & 3 x 3       & 1      & Yes                 & ReLU       & H/8 x W/8 x 512   \\ \hline
\begin{tabular}[c]{@{}c@{}}Upsample\\ Concat(Conv3b)\end{tabular} & -           & -      & No                  & -          & H/4 x W/4 x 512   \\ \hline
Conv7a                                                            & 3 x 3       & 1      & Yes                 & ReLU       & H/4 x W/4 x 256   \\ \hline
Conv7b                                                            & 3 x 3       & 1      & Yes                 & ReLU       & H/4 x W/4 x 256   \\ \hline
\begin{tabular}[c]{@{}c@{}}Upsample\\ Concat(Conv2b)\end{tabular} & -           & -      & No                  & -          & H/2 x W/2 x 256   \\ \hline
Conv8a                                                            & 3 x 3       & 1      & Yes                 & ReLU       & H/2 x W/2 x 128   \\ \hline
Conv8b                                                            & 3 x 3       & 1      & Yes                 & ReLU       & H/2 x W/2 x 128   \\ \hline
\begin{tabular}[c]{@{}c@{}}Upsample\\ Concat(Conv1b)\end{tabular} & -           & -      & No                  & -          & H x W x 128       \\ \hline
Conv9a                                                            & 3 x 3       & 1      & Yes                 & ReLU       & H x W x 64        \\ \hline
Conv9b                                                            & 3 x 3       & 1      & Yes                 & ReLU       & H x W x 64        \\ \hline
Conv10                                                            & 1 x 1       & 1      & No                  & -          & H x W x Classes     \\ \bottomrule
\end{tabular}}
\end{table}

\begin{table}[]
\centering
\caption{Comparison between IOU scores of auto-shot segmenter using one-hot masks vs logit values as annotations.}

\label{tab:logit}
\begin{tabular}{c|ccc}
\bottomrule
Model  & Face & Horse & Car  \\ \hline
One-hot & 82.1 & 69.9  & 70.6 \\ 
Logits & 84.5 & 72.3  & 72.6 \\ \bottomrule
\end{tabular}
\end{table}

\subsection{Various architectures used in GAN-derived Representation Analysis}
In this section, we describe architectures used in GAN-derived representation analysis (Section 5.2). The  small, medium, and large architectures are shown in Table \ref{tab:arch1}, Table \ref{tab:arch2}, and Table \ref{tab:arch3} respectively. 

\subsection{Effects of GAN's Layer Selection}
A study on style mixing from StyleGAN \cite{karras2019style} suggests that information in earlier layers of the GAN's generator controls the higher-level appearance of the output image, whereas late layers control the subtle details. In this experiment, we explore whether choosing different subsets of layers from the generator can affect the performance. 
Similarly to the study in StyleGAN, we roughly split the layers into 3 groups: (A) the coarse style (from resolution $4^2-8^2$), (B) the middle style ($16^2-32^2$), and (C) the fine style ($64^2-1024^2$). Then, we test our one-shot segmenter by feeding different combinations of these groups shown in Table \ref{tab:layer}. The result shows that the representation from group B yields the highest IOU with a slight increase from using all layers, and group A performs the worst. This suggests that the middle layers that control the variation and appearance of facial features are more useful for few-shot face segmentation and that layer selection could play an important role.


\subsection{Arbitrary Segmentation}
\begin{figure}
  \includegraphics[scale=0.56]{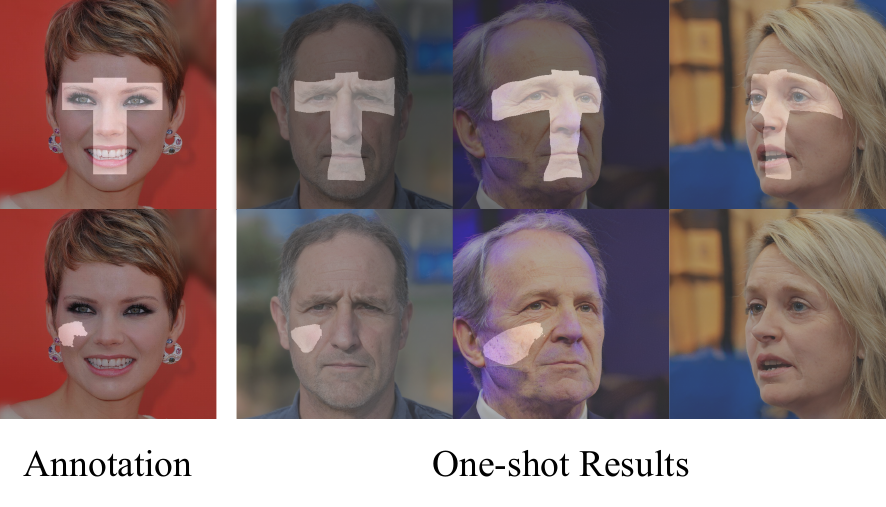}
  \caption{Given segmentation masks with arbitrary, meaningless shapes that have no clear boundary, our approach can infer the same regions across different face images and can even recognize it when the regions are stretched or out of view.}
  \label{fig-weird}
  \vspace{-0.5em}
\end{figure}

We have shown that features from GANs are effective for part segmentation when those parts, so far, correspond to some natural semantic regions such as eyes and mouth. In this experiment, we test whether features from GANs are restricted to those parts and whether our method can generalize to any arbitrary segmented shapes. We manually create random shaped annotations and use them to train our few-shot network. Figure \ref{fig-weird} shows that our method can still handle semantic-less parts and produce consistent segmentation across people and head poses.

\begin{table}[]
\caption{Architecture of S-network.}

\label{tab:arch1}
\resizebox{\columnwidth}{!}{%
\begin{tabular}{|c|c|c|c|c|}
\hline
Layer & Kernel size & Dilation rate & Padding & Output channel size \\ \hline
Conv1 & 3 x 3       & 1             & 1       & 128                 \\ \hline
Conv2 & 3 x 3       & 2             & 2       & 64                  \\ \hline
Conv3 & 3 x 3       & 1             & 1       & 64                  \\ \hline
Conv4 & 3 x 3       & 2             & 2       & 32                  \\ \hline
Conv5 & 3 x 3       & 1             & 1       & number of classes   \\ \hline
\end{tabular}}
\end{table}

\begin{table}[]
\caption{Architecture of M-network.}

\label{tab:arch2}
\resizebox{\columnwidth}{!}{%
\begin{tabular}{|c|c|c|c|c|}
\hline
Layer & Kernel size & Dilation rate & Padding & Output channel size \\ \hline
Conv1 & 3 x 3       & 1             & 1       & 128                 \\ \hline
Conv2 & 3 x 3       & 2             & 2       & 64                  \\ \hline
Conv3 & 3 x 3       & 4             & 4       & 64                  \\ \hline
Conv4 & 3 x 3       & 1             & 1       & 64                  \\ \hline
Conv5 & 3 x 3       & 2             & 2       & 64                  \\ \hline
Conv6 & 3 x 3       & 4             & 4       & 32                  \\ \hline
Conv7 & 3 x 3       & 1             & 1       & number of classes   \\ \hline
\end{tabular}}
\end{table}

\begin{table}[]
\caption{Architecture of L-network.}

\label{tab:arch3}
\resizebox{\columnwidth}{!}{%
\begin{tabular}{|c|c|c|c|c|}
\hline
Layer & Kernel size & Dilation rate & Padding & Output channel size \\ \hline
Conv1 & 3 x 3       & 1             & 1       & 128                 \\ \hline
Conv2 & 3 x 3       & 2             & 2       & 64                  \\ \hline
Conv3 & 3 x 3       & 4             & 4       & 64                  \\ \hline
Conv4 & 3 x 3       & 8             & 8       & 64                  \\ \hline
Conv5 & 3 x 3       & 1             & 1       & 64                  \\ \hline
Conv6 & 3 x 3       & 2             & 2       & 64                  \\ \hline
Conv7 & 3 x 3       & 4             & 4       & 64                  \\ \hline
Conv8 & 3 x 3       & 8             & 8       & 32                  \\ \hline
Conv9 & 3 x 3       & 1             & 1       & number of classes   \\ \hline
\end{tabular}}
\end{table}

\begin{table}[]
\centering
\caption{Comparison of 1-shot segmentation performance with representation from different layers of GANs. }
\label{tab:layer}
\small
\begin{tabular}{ccccccccccccc}
\bottomrule
Layers     & Resolution & weighted mean IOU \\ \hline
A        & $4^2-8^2$        & 59.6          \\
B       & $16^2-32^2$        & \textbf{79.1} \\
C       & $64^2-512^2$         & 69.0          \\
A-B       & $4^2-32^2$        & 75.2          \\
B-C       & $16^2-512^2$        & 75.0          \\
A-B-C (all) & $4^2-512^2$        & \textbf{77.9} \\ \bottomrule
\end{tabular}
\end{table}

\begin{table*}[]
\centering
\caption{Weighted IOU scores on 10-shot face segmentation with representation learned from different tasks / networks.}
\label{fig-huge-table}
\begin{tabular}{c|ccc|ccc}
\bottomrule
Task / Network  & \multicolumn{3}{c|}{4 class} & \multicolumn{3}{c}{10 class} \\ \cline{2-7} 
                 & 1-shot  & 5-shot  & 10-shot  & 1-shot  & 5-shot  & 10-shot  \\ \bottomrule
GANs             & 71.7    & 82.1    & 83.5     & 77.9    & 83.9    & 85.2     \\
VAE              & 55.1    & 69.7    & 72.8     & 51.6    & 58.4    & 65.5     \\
Jigsaw Solving   & 23.3    & 46.3    & 60.0     & 41.6    & 54.9    & 60.4     \\
Colorization     & 32.1    & 39.7    & 51.7     & 49.1    & 55.5    & 66.1     \\
HED              & 38.2    & 48.5    & 60.9     & 48.9    & 67.2    & 70.3     \\
Bilat Filtering  & 10.9    & 22.4    & 49.9     & 29.2    & 45.3    & 54.5     \\ \bottomrule
\end{tabular}
\end{table*}

\begin{figure*}
\centering
  \includegraphics[scale=0.56]{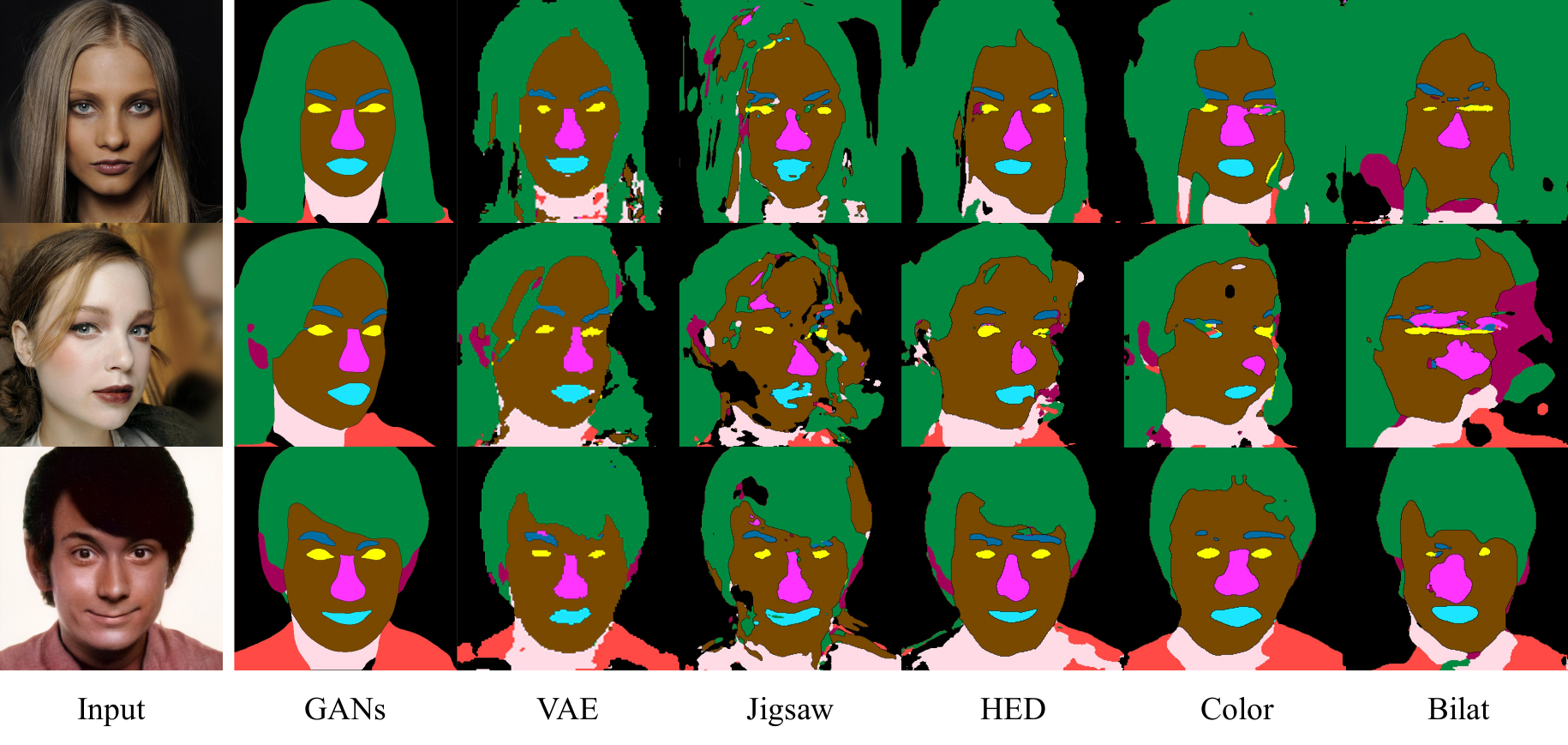}
  \caption{Comparison on 10-shot human face segmentation with representation learned from different tasks / networks.}
  \label{fig-all}
\end{figure*}

\subsection{Comparison on Representation Learned from Other Tasks}
\indent In our paper, we demonstrate the effectiveness of representation from GANs on few-shot semantic part segmentation. However, apart from GANs and generative tasks, there are other tasks and networks that can be used for representation learning. 
In this section, we compare how other representation tasks perform on few-shot human face segmentation. We focus on unsupervised or self-supervised tasks in this study because they require no manual labels and can be applied to any new unseen class. 

We select networks learned to solve 5 different tasks for this experiment: a) VAE \cite{kingma2013auto}, a well-known approach used to synthesize images or find a compact representation of an image through auto encoding with regularized latent distribution, b) jigsaw solving network \cite{noroozi2016unsupervised}, a successful representation learning approach for ImageNet classification that achieves comparable results to a fully-supervised baseline, c) holistically-nested edge detection (HED) network whose nature of the task is closely related to image segmentation, e) colorization network whose representation displays good results in a segmentation task in \cite{larsson2017colorization}, and f) bilateral filtering network which solves a simple task but has to be edge-aware.

\subsection*{Upstream Networks and Feature Extraction}
We train the following baseline networks with human facial images from CelebA dataset \cite{liu2015faceattributes} which comprises 160,000 training images.


\textbf{Variational Autoencoder (VAE)} We use ResNet architecture \cite{he2016deep} for both encoder and decoder and train the network with a perceptual loss from all layers of VGG19 \cite{zhang2018unreasonable}. The generated images are realistic but blurrier compared to those generated from state-of-the-art GANs.

We extract a pixel-wise representation from VAE by feeding an input image into the encoder through the decoder and extracting all activation maps from all convolutional layers in the decoder of VAEs. Then, similarly to GANs feature extraction, all activation maps are upsampled into the dimension of the biggest activation maps and concatenated together in the channel dimension.

\textbf{Jigsaw Solving Network Setup} We follow the setup and network architecture from \cite{noroozi2016unsupervised} to implement a jigsaw solving network. This task asks the network to predict one of the 1,000 predefined permutations of shuffled image patches. To explain the process, first we randomly crop a big square patch from an image and divide the patch into a 3x3 grid. 
All 9 partitioned squares are then cropped again into slightly smaller squares. The 9 squares are then shuffled into one of 1,000 predefined permutations, and the network is trained to predict the 1,000 predefined permutation from the nine squares as input. 

We extract features from all convolutional layers of the jigsaw solving network using the same method as in our VAE representation extraction.

\textbf{Images Translation with Pix2Pix}
we use Pix2Pix \cite{pix2pix2017} framework to create networks that take an image as input and predict some transformed version of that image. We use three types of image transformations: colorization, holistically-nested edge detection (HED) \cite{xie2015holistically}, and bilateral filtering. For colorization, we transform the images to Lab space, then use L channel as input and train the network to predict values in a channel and b channel. For HED, we use HED implementation and pretrained weights from \cite{pytorch-hed}. 

Pix2Pix is a conditional GAN, and a latent optimizer is not needed for embedding the input image because it can take images as input directly to its UNet-based architecture. We feed an input image into Pix2Pix's encoder but only use activation maps from all convolutional layers of the generator (or decoder) to construct a pixel-wise representation.


\subsection*{Results}


\indent The segmentation results are shown in Figure \ref{fig-all} and Table \ref{fig-huge-table}. Representation acquired from GANs produces the best results among all networks. For VAE, the segmentation results are good for 3-class segmentation, second only to GANs. However, the results are noticeably worse in 10-class segmentation because of the bad results in hair class. The segmentation results with representation from the jigsaw solving task can locate facial features, but the quality of contour is poor. 
HED representation has comparatively good segmentation results which could be because segmentation and edge detection problems are closely related, both requiring locating part boundaries. However, since HED often fails to find the nose boundary, nose segmentation is worse than other three previous tasks. Colorization is another task that cannot find the nose boundary as the nose and all facial skin share the same color, and there is no need for the colorization network to learn to discriminate noses from faces. The bilateral filtering task has the worst segmentation results as the network may only learn to find objects' edges and a kernel that can blur images.

\end{document}